%% file: main.tex
\documentclass[12pt]{article}
\RequirePackage[authoryear]{natbib}

\RequirePackage{amsthm,amsmath,amsfonts,amssymb}
\RequirePackage[colorlinks,citecolor=blue,urlcolor=blue]{hyperref}
\RequirePackage{graphicx}
\usepackage[authoryear]{natbib}
\usepackage{float}
\usepackage{enumitem}
\usepackage{url}
\usepackage{bbm}
\usepackage{listings}
\usepackage{booktabs,caption,subcaption}
\usepackage{tikz}
\usetikzlibrary{shapes, arrows.meta, positioning}

\input{commands}

\usepackage[margin=1in]{geometry}

\usepackage{setspace}
\begin{document}

\title{\bf Spatial Neural Network with Transfer Learning} 
\author{Hongjian Yang\\ 
Department of Statistics, North Carolina State University}
\date{}
\maketitle
\thispagestyle{empty}

\newpage
\setcounter{page}{1}

\section{Introduction}

Spatial data is ubiquitous, encompassing a wide range of applications from environmental observations and biological measurements to more recent fields like computer vision. A critical challenge in the analysis of spatial data is spatial prediction, which involves estimating unobserved values based on nearby observations under the assumption of certain correlations. Among parametric algorithms, Kriging is particularly notable ((\cite{kriging})). Described as the best linear unbiased estimator (BLUE), Kriging employs a weighted average of nearby observations, with weights determined by a covariance function typically presumed to be stationary. However, this assumption does not hold in many real-world scenarios, such as data from satellites, monitoring stations, and urban streets, which tend to exhibit non-stationarity (\cite{katzfuss2013bayesian}). Moreover, Kriging faces computational challenges with large datasets, requiring the inversion of the covariance matrix, an operation with a computational complexity of $O(N^3)$ (\cite{chen2020deepkriging}).

A variety of algorithms have been proposed to address the issues highlighted above. \cite{mao2022valid} introduced a model-free method for handling non-stationary spatial data, presenting a significant advancement in the field. Another popular alternative to Kriging is the Vecchia approximation, which simplifies the full Gaussian distribution by conditioning on neighboring values (\cite{vecchia1988estimation}). Drawing on the concept of neighbor-based approximation, \cite{wang2019nearest} developed a nearest neighbor neural network specifically for spatial prediction. Building on these ideas, \cite{chen2020deepkriging} introduced the innovative Deep Kriging framework, which combines spatial basis functions with deep neural networks to model any spatial processes effectively.

While deep learning approaches have demonstrated significant promise in approximating spatial surfaces, they typically require a substantial amount of training data. Transfer learning emerges as an effective solution to this challenge. For instance, \cite{he2019heterogeneous} employed transfer learning to leverage a neural network pre-trained on the ImageNet dataset for Hyperspectral Image Classification (HSI). Similarly, \cite{zhang2012writer} utilized transfer learning to adapt to a common latent representation for writer adaptation. Both the HSI and writer adaptation scenarios, which are limited by small target sample sizes, benefit from the integration of external information from larger datasets.

Spatial applications often grapple with the challenge of small sample sizes. For instance, in estimating PM$_{2.5}$ concentrations, only about 70 high-quality monitoring stations are available throughout North Carolina, and approximately 200 in California (\cite{yang2023data}). Despite this, a large volume of relatively low-quality data exists in these states. Since the spatial distribution of PM$_{2.5}$ tends to be stable, integrating data from these abundant but lower-quality stations could significantly enhance estimation accuracy. Drawing inspiration from the works of \cite{he2019heterogeneous} and \cite{chen2020deepkriging}, this paper proposes a neural network-based transfer learning method that utilizes external information to improve spatial predictions in datasets with limited target data.

\section{Method and Theoretical Properties} 
Let $Y_{i}$ be the observation at spatial location $\bs_i= (s_{i1},s_{i2})$ for $i\in\{1,...,n\}$.  This paper assumes 
\begin{equation}\label{e:Y}
Y_{i} = f(\bs_i;\btheta) + \varepsilon_{i},
\end{equation}
where $f$ is a spatial process that depends on parameters $\btheta$ and $\varepsilon_{i}\iid\mbox{Normal}(0,\tau^2)$ is error.

The spatial process is modelled using a feed-forward neural network (FFNN) with input $\bs_i$.  The FFNN has two stages: in the first stage we deform the spatial coordinates using Radial Basis Function (RBF), and in the second stage the weights are applied to capture the underlying spatial strcture. Below, we describe the model with a single hidden layer in the first stage, and with seven hidden layers of 100 neurons in the second stage.

Following \cite{chen2020deepkriging}, we first use an embedding layer expanding the spatial location into $p$ known basis functions $K_1(\bs),...,K_p(\bs)$.  In particular, we use the Wendland basis function $$\phi(d) = \begin{cases} 
(1 - d)^6(35d^2 + 18d + 3)/3, & d \in [0,1] \\
0, & \text{otherwise}.
\end{cases}$$
where $d$ is the Euclidean distance between observations and knots. Adopting the idea in \cite{nychka2015multiresolution}, we use a multi-resolution with the knots arranged on a rectangular grid. In particular, we used four level of resolutions, and at each level, let $u_i, i = 1,2,3,4$ be a rectangular grid of points. The basis function is defined as $\phi^*(\bs) = \phi(d) = \phi(||\bs - u_i||)$. After the first stage, we have a basis function representation of $\mathbf{x} \in \mathbb{R}^{139}$. 

The second stage of the neural network is defined as follows:
\begin{align*}
\mathbf{h}_1 &= \text{ReLU}(\mathbf{W}_1 \mathbf{x} + \mathbf{b}_1) \\
\vdots \\
\mathbf{h}_7 &= \text{ReLU}(\mathbf{W}_7 \mathbf{h}_6 + \mathbf{b}_7) \\
y &= \mathbf{W}_8 \mathbf{h}_3 + b_8
\end{align*}

where:
\begin{itemize}
    \item $\mathbf{W}_1 \in \mathbb{R}^{100 \times 139}, \mathbf{b}_1 \in \mathbb{R}^{100}$ are the weights and biases of the first hidden layer.
    \item $\mathbf{W}_2 \hdots \mathbf{W}_7 \in \mathbb{R}^{100 \times 100}, \mathbf{b}_2 \hdots \mathbf{b}_7 \in \mathbb{R}^{100}$ are the weights and biases of the second and third hidden layers.
    \item $\mathbf{W}_8 \in \mathbb{R}^{1 \times 100}, b_8 \in \mathbb{R}$ are the weights and bias of the output layer.
    \item $\text{ReLU}(\cdot)$ is the Rectified Linear Unit activation function.
    \item $y \in \mathbb{R}$ is the output of the network.
\end{itemize}

A visual illustration of the architecture of the two-stage neural network are provided in the supplementary materials.

The neural network above is first trained on a large external data, and all parameters are transferred to the target data set, since we assume the distribution of the covariates $X$ are the same. 

Many recent advancement of transfer learning in NLP and Computer Vision area shows that the method proposed could improve the estimation performance. For example, Segment Anything Model (\cite{sam}) learns from a millions of images and is the state-of-the-art in segmentation task. The basic idea behind both Segment Anything Model and the proposed approach is that the neural network is learning latent representation of the training feature, and can decode the learned feature and adapt to new tasks quickly.

\section{Simulation}
To evaluate the performance of the proposed method above, a simulation study is carried out using both stationary and non-stationary data on a unit square $[0, 1]^2$. 

The stationary data is a Matern spatial process as defined by \cite{SPDE}, where the correlation between two spatial location $\bs_{i}, \bs_{j}$ is 
$$C(\bs_{i}, \bs_{j}) = \frac{\sigma^2}{2^{\nu - 1} \Gamma(\nu)} (\kappa||\bs_{i} - \bs_{j}||)^\nu K_\nu(\kappa||\bs_{i} - \bs_{j}||)
$$
where $K_\nu$ is the modified Bessel function of the second kind with smoothness factor $\nu = 1$, and $\kappa = \frac{\sqrt{8}}{\rho}$ where $\rho = 0.2$ is the spatial range, and $\sigma^2 = 1$ is the spatial range. The nugget error $\epsilon_i = 0.01$ is the i.i.d noise.

\begin{figure}[h!]
    \centering
    \includegraphics[scale=0.4]{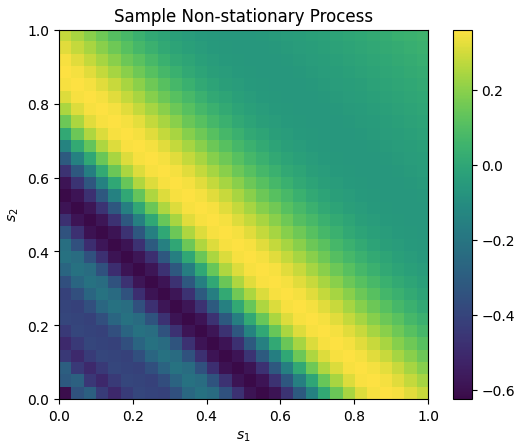}
    \includegraphics[scale=0.4]{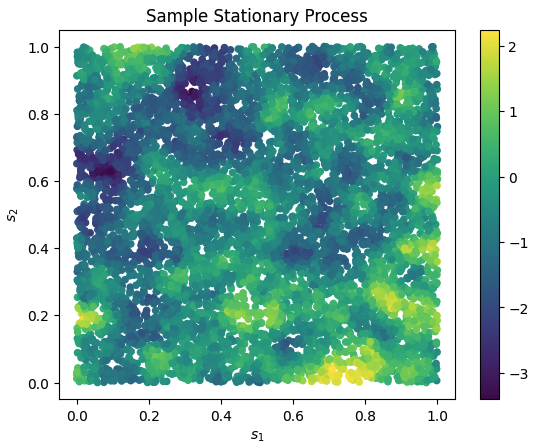}
    \caption{Simulation data}
    \label{fig:spatial_process}
\end{figure}

The non-stationary process is defined in \cite{chen2020deepkriging} with $Y_i = sin\{30(\Bar{\bs_i}-0.9)^4\}cos\{2(\Bar{\bs_i}-0.9)\}+(\Bar{\bs_i}-0.9)/2$, where $\bs_i = (\bs_{i1}, \bs_{i2})$ and $\Bar{\bs_i} = \frac{\bs_{i1} + \bs_{i2}}{2}$. To distinguish these two processes, in this paper we define the stationary process as $Y_{iS}$ and the non-stationary process as $Y_{iN}$. An independent nugget variance of $\nu^2 = 1e^{-6}$ is added to the non-stationary process. Figure~\ref{fig:spatial_process} shows the spatial surface of sample realizations.

\newpage

\begin{figure}[h!]
    \centering
    \includegraphics[scale=0.4]{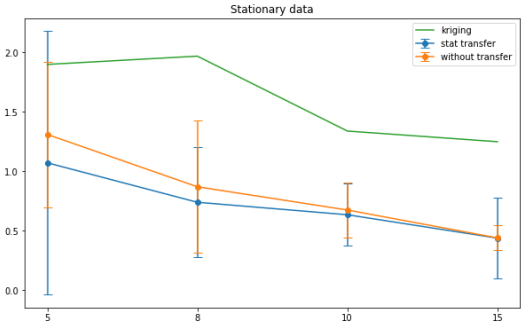}
    \caption{Stationary process MSE}
    \label{fig:stat}
\end{figure}

\begin{figure}[h!]
    \centering
    \includegraphics[scale=0.4]{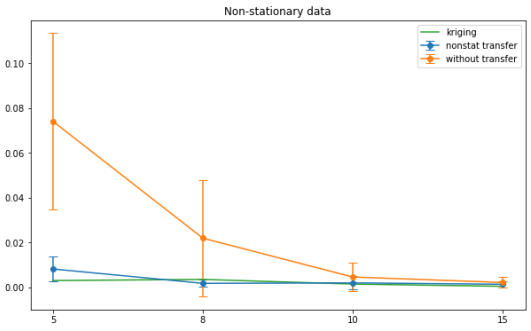}
    \caption{Non-stationary process MSE}
    \label{fig:nonstat}
\end{figure}

For each process, the source data set each consists of $N = 4900$ observations. We evaluate the performance on the target data set of $N = 25, 64, 100, 225$ observations. Figure~\ref{fig:stat} displays the MSE comparison of 1). source data pre-trained MSE on target data 2). target data set only, and 3). Kriging result on stationary data. Figure~\ref{fig:nonstat} compares the results on non-stationary data.

During the pre-training stage on external data, a total of 1500 epochs are used with a learning rate of 0.001. A trace plot with validation set is monitored to make sure the neural network has converged. During the tuning stage on target data, all parameters from the pre-training stage are updated. A total of 1000 epochs with a learning rate of 0.001 is used. 

\section{Conclusion}
As we can see from the figures above, for stationary data, the proposed approach outperforms both target only setting and the traditional Kriging approach. As target set gets larger, the proposed method and target only neural network converges. 

For the non-stationary scenario, the proposed method significantly outperforms target only approach when the target sample size is less than 100, and similar to the stationary case, these two neural network converges when the sample size gets larger. 

The proposed transfer learning approach is a first step towards spatial transfer learning. Following \cite{he2019heterogeneous}, we tune all parameters in the model. One possible future extension is to add additional layers to the neural network, fix the first seven layers in the network, and only tune additional layers. Also, the fully connected neural network may not capture the spatial dependence well. It is possible to combine the 4N network proposed by \cite{wang2019nearest} or the graph neural network \cite{klemmer2023positional}. 

\newpage

\begin{singlespace}
	\bibliographystyle{rss}
	\bibliography{refs}
\end{singlespace}

\section{Supplementary Materials}
\subsection{Neural Network Architecture}
\begin{tikzpicture}[node distance=0.5cm]
  \tikzstyle{neuron}=[circle,draw=black,minimum size=17pt,inner sep=0pt]
  \tikzstyle{input neuron}=[neuron, fill=green!50];
  \tikzstyle{output neuron}=[neuron, fill=red!50];
  \tikzstyle{hidden neuron}=[neuron, fill=blue!50];
  \tikzstyle{hidden neuron fc}=[neuron, fill=orange!50];
  \tikzstyle{annot} = [text width=4em, text centered]

  \foreach \name / \y in {1,...,2}
    \node[input neuron] (I-\name) at (0,-\y) {};

  \foreach \name / \y in {1,...,5}
    \path[yshift=1.5cm]
      node[hidden neuron] (H-\name) at (2.5cm,-\y cm) {};

  \foreach \name / \y in {1,...,3}
    \path[yshift=1cm]
      node[hidden neuron fc] (F-\name) at (5cm,-\y cm - 1cm) {};

  \node[output neuron,right of=F-2] (O) {};

  \foreach \source in {1,...,2}
    \foreach \dest in {1,...,5}
      \path (I-\source) edge (H-\dest);

  \foreach \source in {1,...,5}
    \foreach \dest in {1,...,3}
      \path (H-\source) edge (F-\dest);

  \foreach \source in {1,...,3}
    \path (F-\source) edge (O);

  \draw[dashed, red, thick] (3.75, -5.5) -- (3.75, 0.5);

  \node[annot,above of=H-1, node distance=2cm] (hl) {{\color{red}RBF layer}};
  \hspace{2cm}
  \node[annot,right of=hl] {{\color{blue}Fully connected seven layer}};
\end{tikzpicture}

\end{document}

%% file: commands.tex
\newcommand{\btheta}{ \mbox{\boldmath $\theta$}}

\newcommand{\bs}{ \mbox{\bf s}}

\newcommand{\iid}{\stackrel{iid}{\sim}}

\newcommand{\beq}{ \begin{equation}}
\newcommand{\eeq}{ \end{equation}}
\newcommand{\beqn}{ \begin{eqnarray}}
\newcommand{\eeqn}{ \end{eqnarray}}

\usepackage{caption}